# What's Race Got to do with it?
# Predicting Youth Depression Across Racial Groups Using Machine and Deep Learning

Nathan Zhong
River Hill High School
Clarksville, MD, USA
NathanYZhong@gmail.com

Nikhil Yadav
Division of Computer Science, Mathematics and Science
St. John's University
Queens, NY, USA
yadavn@stjohns.edu

*Abstract*— Depression is a common yet serious mental disorder that affects millions of U.S. high schoolers every year. Still, accurate diagnosis and early detection remain significant challenges. In the field of public health, research shows that neural networks produce promising results in identifying other diseases such as cancer and HIV. This study proposes a similar approach, utilizing machine learning (ML) and artificial neural network (ANN) models to classify depression in a student. Additionally, the study highlights the differences in relevant factors for race subgroups and advocates the need for more extensive and diverse datasets. The models train on nationwide Youth Risk Behavior Surveillance System (YRBSS) survey data, in which the most relevant factors of depression are found with statistical analysis. The survey data is a structured dataset with 15000 entries including three race subsets each consisting of 900 entries. For classification, the research problem is modeled as a supervised learning binary classification problem. Factors relevant to depression for different racial subgroups are also identified. The ML and ANN models are trained on the entire dataset followed by different race subsets to classify whether an individual has depression. The ANN model achieves the highest F1 score of 82.90% while the best-performing machine learning model, support vector machines (SVM), achieves a score of 81.90%. This study reveals that different parameters are more valuable for modeling depression across diverse racial groups and furthers research regarding American youth depression.

## I. Introduction

Depression, or Major Depressive Disorder (MDD), is a mental illness and disorder of the brain which engulfs individuals in a prolonged state of sadness and hopelessness [1]. The symptoms of depression include problems with sleep, appetite, cognitive abilities, and motivation, often disrupting school, work, or home life. Depression is a significant health problem worldwide and affects people of all ages. In the U.S., the average anxiety and depression rate is 32.3% [2] and over 16% of youth aged 12 to 17 had at least one major depressive episode this past year [3]. Moreover, depression is a leading factor in suicide, where 60% of people who die by suicide had depression or other disorders [4]. The severe nature of depression makes detection and intervention crucial as it can have a profound impact on our communities.

Many factors contribute to depression, often a mix of biological, psychological, and external factors like life experiences and stress. In addition, many people with depression do not report symptoms, making it hard to detect and treat [5]. Especially among high school students, depression is a hard topic to discuss, and they are often ashamed to reveal signs of depression. In the United States, the teenage mental health crisis is extreme, with the percentage of American students saying they feel "persistent feelings of sadness or hopelessness" shooting from 26% to 44% from 2009 to 2021 [6]. The COVID-19 pandemic and the rise of social media also add to the severity of the teenage depression crisis [7]. Therefore, it is vitally important for policymakers and health officials to find a reliable technique to accurately identify occurrences of the disorder.

Researchers attempt to identify depression utilizing ML and neural network models. Some ML approaches train data from social media sites such as Twitter. Govindasamy and Palanichamy use the Naïve Bayes and NBTree ML models on public Twitter data to classify depression, achieving up to 97% accuracy [8]. These ML models often focus on people of all ages and rely exclusively on public social media data. Haque et al., focuses on youth depression by using hundreds of categorical variables from Youth Minds Matter data (Australian) as predictors in ML models, achieving up to 95% accuracy [9]. A study by Lee et al. used both machine learning models and ANNs to predict depression, analyzing national survey data for American adults with hypertension. The models achieved accuracies of around the 80s for the five conventional ML models, while the highest Area Under the Curve Value (AUC), a measure of ability to distinguish between positive and negative classes, was achieved with ANNs at 81.3% [10].

Deep learning has become a popular method for identifying many mental health issues. Deep learning models consist of networks of neurons, containing interconnected nodes organized into layers, similar in structure to the human brain. Some types of networks are ANNs, convolutional neural networks (CNNs), and natural language processing (NLP). ANNs are suited for dealing with tabular data like survey responses, while CNNs specialize in image processing and can be useful for examining medical patients. NLPs can track sentiments from text and even monitor the behavior of social media users.

Many recent health-related papers make use of deep learning methods. Some researchers use CNNs to analyze physical features and detect signs of illness. Acharya et al. use CNNs on facial features to identify depression with a precision of up to 95% [4]. Geraci et al. applied NLP to the phenotyping of psychiatric diagnosis for youth, identifying depression with lower precision around 80% [11]. In a study by Allahyari and Roustaei, an ANN model was trained using questionnaires for the adults in Birjand and Mashhad, reaching a high accuracy of 99.2%. Among the most important predictors included education, employment status, and age [12].

Overall, few papers attempt to utilize neural networks to examine depression across different racial groups or among the American high school youth population. This study utilizes both conventional ML models and ANNs and is based on the nationwide YRBSS survey data. First, the four most statistically significant factors related to youth depression (risk factors from the survey) are calculated for the entire dataset. Next, ML and ANN models are built using the factors as predictors. The ML models are then compared with the ANNs using F1 score as the objective comparison statistic. The study provides a racial lens by exploring the different factors that affect depression in different racial groups and fitting ANNs on race subset data.

The remainder of this paper is structured as follows: Section II outlines the methodology of statistical analysis to prepare training data and creates the ML and ANN models. Section III includes F1 scores obtained by the models, extends ANNs to racial subsets, and presents findings. Finally, section IV discusses the potential uses of this research and envisages future work based on the study results.

## II. METHODOLOGY

The overall methodology of the paper is represented in Fig. 1 below. Data was collected and converted to comma separated values (csv) from the YRBSS survey dataset published online [13]. Three race subset csv files, *fully Asian, Latino/Hispanic, fully black*, were created from the dataset. Statistical significance tests were used to select, validate, and compare the most relevant risk factors (features) for depression across all datasets. Conventional ML models and a parameter-tuned ANN were trained for each of the datasets with F1-scores from each of these models evaluated and compared.

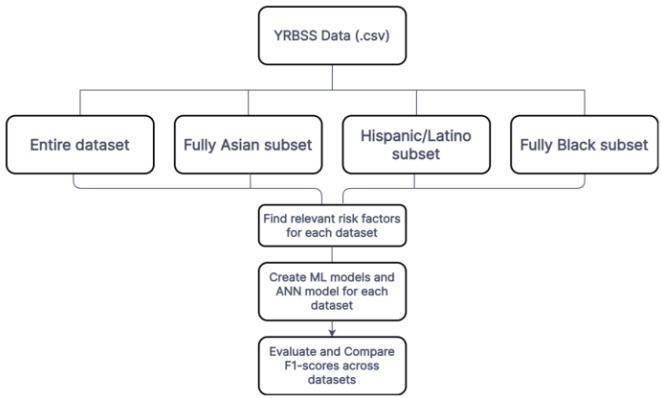

FIG. 1. EXPERIMENTAL PROTOCOL

### A. Data Collection

The YRBSS survey is a medium-high granularity dataset from the CDC. The YRBSS handbook was used for reference. The survey was conducted in late 2021 and provides data representative of 9$^{th}$ through 12$^{th}$ grade students in public and private schools in all 50 states and the District of Columbia [13]. The survey is conducted biennially with the next survey to come in 2023 after the fall semester.

The raw data consists of 17 rows and 149 columns. Each column has a label. Some labels are a survey question number, for example: q5 from the handbook states: *what is your race*? Other labels are additional information, for example: qnfrcig represents the percentage of students who smoked cigarettes frequently. Each row of the csv represents an individual's responses to the survey questions and statistics from the additional information labels. Most responses are coded numerically, for example the answers to yes or no questions responses were coded to 1 or 2. The only responses that were categorical were the ones in the q5 column: race. This was coded as one or more letters depending on if a person is mixed.

The raw data was then preprocessed. Rows with no race entries were deleted (empty q5 column), leaving 15433 entries. The additional information columns were deleted, leaving 99 columns (the 99 survey questions). This resulting dataset is referred to as the *All Races* dataset.

Next, using the excel filter method on q5, subsets of Fully Asian and Fully Black were created by singling out the "B" responses for fully Asian and singling out the "C" responses for fully Black. Other race responses include "E" for White and "A" for American Indian or Alaskan Native, shown in Table. 1. Finally, a Latino/Hispanic subset was created: the *All Races* csv was filtered through the q4 column, which asks, a*re you Hispanic or Latino?*

In the datasets, the data in q25 was identified as the dependent variable. The question states: *During the past 12 months, did you ever feel so sad or hopeless almost every day for two weeks or more in a row that you stopped doing some usual activities*? essentially asking if the student had depression, which is the basis of this study. The best predictor variables (independent variables, or features) are found in section III. A.

TABLE 1. EXCERPT FROM TWO ENTRIES OF THE *ALL RACES* CSV

| Data | | | | | | |
|---|---|---|---|---|---|---|
| *q2* | *q3* | *q4* | *q5* | *q6* | ... | *qn99* |
| 2 | 2 | 1 | B E | 1.73 | | 1 |
| 1 | 2 | 1 | A | 1.63 | | 1 |

### B. Software Tools

Data preprocessing, statistical analysis, and ML and ANN model training were performed with code written in Python using the Pandas, NumPy, Scikit-learn, SciPy, Keras, and TensorFlow libraries.

### C. Feature Selection

Two statistical significance tests, one proportion Z-test and p-value test, were performed on the data to select and validate the most relevant factors contributing to depression for each of the datasets.

The Z and p-value tests were calculated for the *All Races* csv. First, the proportion of people of all races with depression was found: 0.39597.

Next, the Z-test was run by looping through each column. The proportion of people with certain responses to a question who had depression was compared with 0.39597 using the proportions_ztest function built into SciPy. The function returned Z-scores for each column, calculated as shown on the next page where x represents the observed value, μ represents mean, and σ represents standard deviation:

$$Z = \frac{x - \mu}{\sigma}$$

Fundamentally, a high magnitude Z-score means that the column question responses have a great effect on whether the individual will have depression or not, showing that the question is statistically significant. Columns with Z-score higher or lower than the Z statistics on the Z-score table, signifying statistical significance to depression, were identified. The four columns with highest magnitude Z-scores are displayed in Table 2. These were chosen to avoid overfitting observed when more than four features were used in the models.

Similarly, p values were calculated for the four columns to validate their statistical significance to depression. The p values were well below 0.05, proving that the four features for each of the datasets were statistically significant to be used as predictor variables.

The same process was completed for the three race subsets. The results are shown in Table 2.

From each dataset, four new csv files were created with five columns each. For each file, the first four columns were the most relevant columns found above. The last column was the q25 column coded to binary (0 or 1). In this format, the csv files were ready for ML and deep learning modeling.

TABLE 2. TOP 4 FEATURES PER RACE, RANKED BY Z-TEST P-VALUE

| All Races | | | |
|---|---|---|---|
| Feature | Z score | p value | p < 0.05 |
| q2 (sex) | 22.705 | 4.03E-114 | Yes |
| q23 (bullied at school) | 28.573 | 1.46E-179 | Yes |
| q24 (e- bullied) | 32.819 | 3.12E-236 | Yes |
| q34 (tried vape) | 27.458 | 5.59E-166 | Yes |
| **Fully Asian** | | | |
| Feature | Z score | p value | p < 0.05 |
| q16 (physical fight) | 5.046 | 4.50E-07 | Yes |
| q24 (e- bullied) | 5.966 | 2.44E-09 | Yes |
| q30 (tried cigarette smoking) | 6.981 | 2.93E-12 | Yes |
| q34 (tried vape) | 5.582 | 2.38E-08 | Yes |
| **Latino/Hispanic** | | | |
| Feature | Z score | p value | p < 0.05 |
| q2 (sex) | 13.283 | 2.90E-40 | Yes |
| q18 (seen violence in neighborhood) | 10.734 | 7.04E-27 | Yes |
| q24 (e- bullied) | 14.37 | 7.94E-47 | Yes |
| q34 (tried vape) | 15.132 | 9.91E-52 | Yes |
| **Fully Black** | | | |
| Feature | Z score | p value | p < 0.05 |
| Q15 (threatened with weapon at school) | 5.993 | 2.06E-09 | Yes |
| Q23 (bullied at school) | 8.7 | 3.32E-18 | Yes |
| q24 (e- bullied) | 10.379 | 3.09E-25 | Yes |
| q34 (tried vape) | 8.624 | 6.47E-18 | Yes |

*D. F1 Scores*

The objective comparison statistic and measure of results for the upcoming models was the F1 score, a measure of a binary classification model's predictive performance. The score is defined as the harmonic mean of the recall and precision, and the formula is given below. Precision is the fraction of the number of positive predictions correctly classified to the total number of predictions classified as positive. Recall:

$$F_1 = \frac{2 \times (\text{precision} \times \text{recall})}{\text{precision} + \text{recall}}$$

F1 score is used because it accounts for both false positives and false negatives and still relays true model performance when the dataset is imbalanced. Around 30%-40% of the depression class was observed in the dataset.

*E. Conventional ML Models*

The following conventional ML models were utilized to predict the occurrence of depression across the datasets:

- Support Vector Machine (SVM), linear basis function kernel
- Logistic Regression
- Decision Tree, visualized in Fig. 2.
- Random forest

These conventional models are prevalent in research regarding mental health issues. In Lee et al.'s study on depression, random forest and linear SVM were utilized on tabular data like this study [10]. Rois et al. use polynomial SVM, decision tree models, and Logistic Regression models to predict stress [14].

The research problem was modeled as a binary classification problem because of the binary and discrete nature of depression and the variety of ML models available for supervised learning and classification. The training data was created using both a hold-out method with 75/25 training testing split and cross validation with 10 folds across all csv files. A row of training data for the *All Races* csv is shown Table 3. The conventional ML models were fitted onto the training data using the first four columns as predictor variables and the last column as the prediction variable, creating 32 unique models.

The purpose of the ML models was to lay a benchmark to contrast with the deep learning-based ANN model.

Fig. 2. One decision tree of random forest model for Asian subset

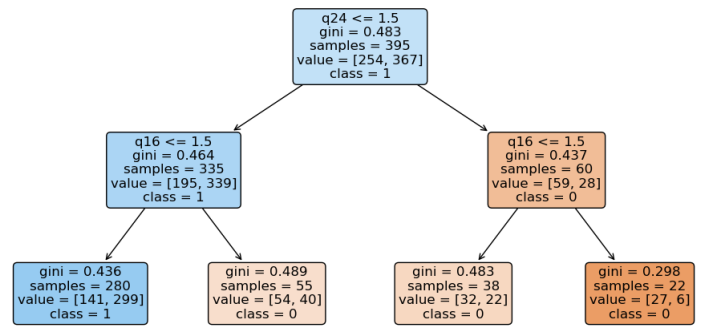

TABLE 3. ALL RACES CSV TRAINING DATA ROW

| Predictor variables (gender, being bullied, being electronically bullied, vaped) | | | | Prediction variable (depressed) |
|---|---|---|---|---|
| q2 | q23 | q25 | q34 | q25 |
| 1 | 2 | 1 | 1 | 0 |

### F. ANN

ANN models were used to predict depression alongside the ML models. The ANN models were trained the same way as the ML models, using the first four columns as predictors and the last column as predictions. Both the holdout and cross validation methods were used, resulting in eight unique models.

Multi-layer perceptrons were used to train the models to classify inputs to the target of depression. The Adam optimizer was used in each model because of its popularity and adaptive learning rate. Each ANN model had three layers: the input, hidden, and output layer with the input layer being four dimensional (one dimension per predictor/feature). One hidden layer was chosen based on Zhao's study, which has a neural network with a similar configuration but with 11 features [15].

### G. ANN Parameter Tuning

Hyperparameter tuning to improve and optimize F1 score was explored. A function was created to loop through 1-40 neurons, nine different activation functions, and 5-20 epochs due to the low dimensions of the models and size of the dataset. For each model, the optimal number of epochs was found to be around 9 and the optimal activation function was found to be sigmoid for all layers.

The number of neurons in the hidden layer was around 24 for all the ANN models, which is over double the dimension of the input layer, a sign of possible overfitting [16]. To find a more economical number of neurons for the hidden layers, models looped through 1-8 neurons in the hidden layer. These models had 6-8 neurons and obtained F1 scores fractionally lower than the original models which require significantly more computational resources to build.

For example, using holdout on the *All Races* dataset, 7 was the number of neurons from 1-8 that achieved the highest F1 score. 7 neurons achieved 0.7804 while 23 neurons achieved 0.7821, which is slightly higher. The neural network configuration in Fig. 3 was decided upon as a result.

FIG. 3. ALL RACES ANN MODEL VISUALIZATION

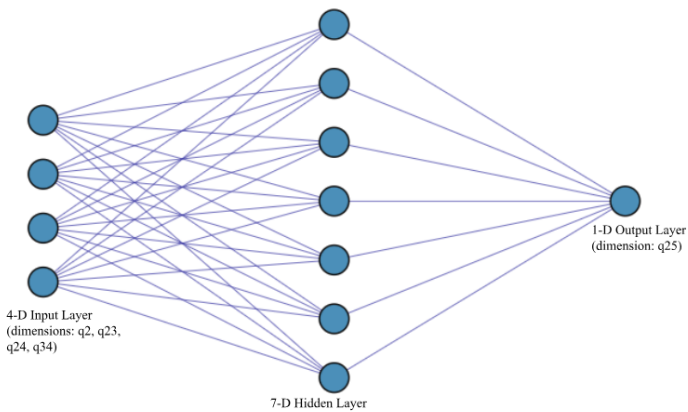

## III. RESULTS

### A. Statistical Analysis, F1 Overview

As referred to Table 2, it was found that sex (gender), being bullied both at school and electronically, and using vape were significant depression predicting factors/features for the *All Races* dataset. For other race groups, some other features were also relevant. For Asians, being in a physical fight and cigarette smoking were significant features for depression. For Latino/Hispanics, witnessing violence in the neighborhood was a significant feature. For Blacks, being threatened with a weapon at school was a significant feature.

Using significant features as mentioned above, the F1 scores for each model were calculated, shown in Fig. 5 and Table 4. Using both holdout and cross validation methods, five model types were run for each race subset – SVM, Logistic Regression, Decision Tree, Random Forest, and ANN – for 40 total scores.

FIG. 4. CROSS VALIDATION (TOP) AND HOLD-OUT (BOTTOM) F1 SCORES

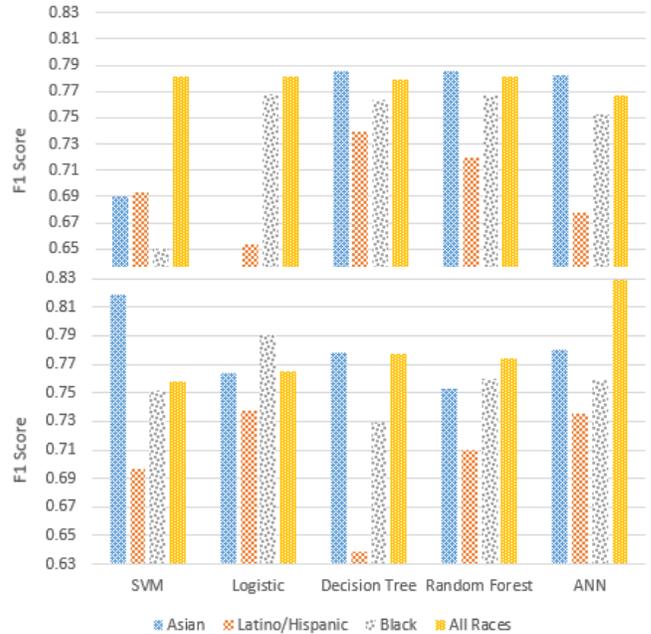

TABLE 4. F1 SCORES

|  | SVM | Logistic Reg. | Decision Tree | Random Forest | ANN |
|---|---|---|---|---|---|
| **Asian** | 0.6903 | 0.6354 | 0.7859 | 0.7857 | 0.7822 |
|  | 0.819 | 0.764 | 0.778 | 0.7523 | 0.7804 |
| **Latino/ Hispanic** | 0.6929 | 0.6542 | 0.7398 | 0.72 | 0.6775 |
|  | 0.6964 | 0.7378 | 0.6385 | 0.71 | 0.7354 |
| **Black** | 0.6508 | 0.7683 | 0.7633 | 0.7672 | 0.7525 |
|  | 0.7522 | 0.79 | 0.7293 | 0.76 | 0.7592 |
| **All Races** | 0.7814 | 0.781 | 0.7796 | 0.7814 | 0.7673 |
|  | 0.7577 | 0.7655 | 0.7775 | 0.7737 | 0.829 |

The four-feature combination per race group predicted depression with varying precision. For the Latino/Hispanic group, the results for all 10 models were low, with a mean of 0.7003. The Fully Asian, Black, and All Races groups had average F1 scores of 0.7573, 0.7493, 0.7794.

### B. Conventional ML Models

The conventional ML models obtained F1 scores ranging from 0.65 to 0.82. Training and testing using the hold-out method produced a higher average score compared to cross validation by almost 0.02 (0.7568 and 0.7378). Asian Subset SVM using hold-out had the highest score out of all 32 models at 0.819. No other ML models achieved results greater than 0.80.

The use of the cross validation method on the three race subsets (the minority group as a whole) was analyzed across the four conventional models. Using SVM, the average F1 score for minorities was (0.6903+0.6929+0.6508)/3 = 0.6780. Logistic Regression, Decision Tree, and Random Forest obtained 0.686, 0.763, and 0.7576, respectively. Next, hold-out models were analyzed for the minorities. SVM, Logistic Regression, Decision Tree, and Random Forest obtained 0.7559, 0.7639, 0.7153, 0.7408, respectively.

Using cross validation for the *All Races* dataset, the four ML models produced F1 scores of 0.7814, 0.781, 0.7796, 0.7814 in the order above. With the hold-out method, scores were 0.7577, 0.7655, 0.7775, 0.7737.

The ML models had adequate and similar F1 scores. It is notable that Decision Tree and Random Forest often obtain scores over 0.75 for all four race datasets. SVM and Logistic Regression are weak when used on the minority group with cross validation, often obtaining a score lower than 0.70.

### C. ANN Models

As mentioned in section II. F, when using ANN models, the number of optimal epochs for each subset was around 9. A figure is shown below for the Asian subset. The F1 score peaked at epoch 10 here.

FIG. 5. F1 SCORES OVER EPOCHS

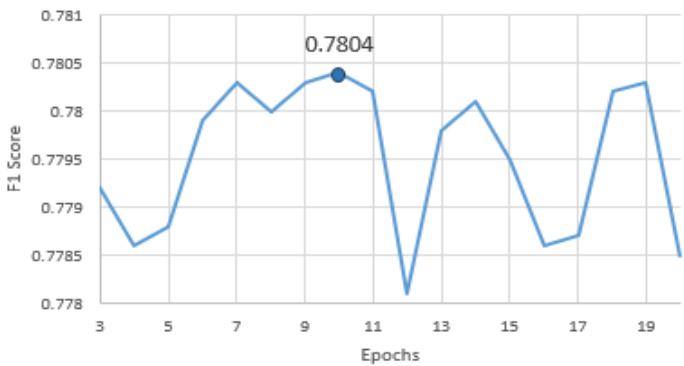

For the minority group, cross validation results on the ANN model were mediocre, obtaining an F1 score of (0.7822+0.6775+0.7525)/3 = 0.7374. The 0.6775 was from the Latino/Hispanic subset. Hold-out on the minority group gives a score of 0.758, close to the SVM and Logistic models using the same testing training split.

On the *All Races* dataset using cross validation, the ANN model performed poorer than all the conventional models, obtaining a score of just 0.7673. Using hold-out on All Races, the ANN model performed better than all the ML models, obtaining 0.829, the first and only result above 0.82. Overall, the ANN models produced consistent F1 scores except in the Latino/Hispanic subset, where the score dipped below 0.75 for both cross validation and hold-out. Excluding the two Latino/Hispanic cases, all results for ANN models fall between 0.75 and 0.80, except for the 0.829 peak.

ANN and Random Forest have relatively consistent results when measured by the standard deviation of all 8 models. ANN models achieve standard deviation of F1 score of 0.0406, while SVM, Logistic Regression, Decision Tree, and Random Forest obtain standard deviations of 0.0527, 0.0553, 0.0459, 0.0260. Random Forest has the most consistent results, but its mean score over the 8 models is 0.7563 while the mean for ANN models is 0.7604. The mean scores for the other models are all under 0.75.

### D. Discussion of Results

All four race groups had the common features of being electronically bullied and using vape as two of the four most significant signs for depression. Other features varied as mentioned in section III. A - different race groups exhibited different significant features for depression.

ANN, Decision Tree, and Random Forest models all obtained relatively consistent F1 scores that rarely dropped below 0.70 while SVM and Logistic Regression were less consistent, obtaining many sub-scores.

ANN did not have a trend of outperforming other models, only having the strongest F1 score for the *All Races* dataset with the holdout method. This is expected because the model was built on more voluminous data without consideration about relevant features prominent in different race groups.

Random Forest approach yielded strong results compared to other conventional models. In the study, Random Forest achieved the highest F1 score with the lowest standard deviation across the 8 models. Also, Random Forest works well with many types of inputs and requires no data normalization [17] and the computational cost and training time is low, so it could work well for diverse datasets irrespective of size.

### IV. CONCLUSION

This paper highlighted different features related to depression in different racial groups and compared the effectiveness of using conventional ML models and ANNs to predict depression given tabular survey data.

This study prompts the use of more extensive and diverse datasets, preferably across different countries and age ranges. Questions that could be explored are: *How do significant features related to depression change across geographic location? How do these features change as students transition to college and beyond?* The answers to these questions could be compared with findings from social science research for a more well-rounded and concrete understanding of depression. Mental health experts and public health policymakers could utilize this information to better identify and treat depression cases across different racial groups across regions.

The adequate F1 scores of both ML and ANN models confirm the ability of machine and deep learning to diagnose mental health issues, and the success of neural networks with the large *All Races* dataset demonstrates neural networks' potential to pinpoint depression accurately in an individual, especially with larger datasets to train from. The deep learning approach could be investigated in future work as more tabular youth depression data is published. ANN models have potential to perform better, since conventional ML models can degrade in performance with large quantities of data, while ANNs can continually achieve better performance as data becomes more available and voluminous [18].

This study can be adapted so that the methods remain the same but instead of depression, suicide risk is being analyzed. Also, the results of the study confirm differences exist between factors contributing to depression for different racial groups [19] and advocates for race-based diagnosis of depression [20].

Soon, YRBSS will conduct the 2023 survey. The same methods could be run on the new dataset. In addition to the ML and ANN models, a time series model could be experimented on with the aim of identifying any trends in significant features among minority groups.